%% file: main.tex
\documentclass[letterpaper, 10 pt, conference]{ieeeconf}  

\usepackage{graphics} 
\usepackage{graphicx}
\usepackage{subfig}
\usepackage{multirow}
\usepackage{array}
\usepackage{booktabs}
\input{mathstuff}

\usepackage{cite}
\usepackage{xcolor}
\usepackage{amsmath}
\usepackage{amssymb}
\usepackage{enumitem}

\newcommand{\ours}{GMM-Det}
\newcommand\blfootnote[1]{%
  \begingroup
  \renewcommand\thefootnote{}\footnote{#1}%
  \addtocounter{footnote}{-1}%
  \endgroup
}
\begin{document}

\title{Uncertainty for Identifying Open-Set Errors in Visual Object Detection}

\author{Dimity Miller, Niko S\"underhauf, Michael Milford, and Feras Dayoub}

\maketitle
\blfootnote{The authors are with Queensland University of Technology (QUT), Brisbane, Australia.Dimity Miller is also affiliated with the CSIRO Robotics and Autonomous Systems Group. This research has been conducted by the Australian Research Council (ARC) Centre of Excellence for Robotic Vision (Grant CE140100016) and supported by the QUT Centre for Robotics. Dimity Miller acknowledges continued support from the CSIRO Machine Learning and Artificial Intelligence Future Science Platform. Contact d24.miller@qut.edu.au and code is made available by the authors at github.com/dimitymiller/openset\_detection}

\begin{abstract}Deployed into an open world, object detectors are prone to open-set errors, false positive detections of object classes not present in the training dataset. We propose \ours{}, a real-time method for extracting epistemic uncertainty from object detectors to identify and reject open-set errors. \ours{} trains the detector to produce a structured logit space that is modelled with class-specific Gaussian Mixture Models. At test time, open-set errors are identified by their low log-probability under all Gaussian Mixture Models. We test two common detector architectures, Faster R-CNN and RetinaNet, across three varied datasets spanning robotics and computer vision. Our results show that \ours{} consistently outperforms existing uncertainty techniques for identifying and rejecting open-set detections, especially at the low-error-rate operating point required for safety-critical applications. \ours{} maintains object detection performance, and introduces only minimal computational overhead. We also introduce a methodology for converting existing object detection datasets into specific \emph{open-set} datasets to evaluate open-set performance in object detection.

\end{abstract}

\input{sections/Introduction}

\input{sections/RelatedWork}

\input{sections/Method}

\input{sections/Evaluation}

\input{sections/Results}

\input{sections/Conclusion}

\bibliographystyle{IEEEtran}
\bibliography{refs}

\end{document}

%% file: mathstuff.tex
\newcommand{\vect}[1]{\mathbf{ #1}}
\newcommand{\vectg}[1]{{\boldsymbol{ #1}}}

\newcommand{\T}{^\mathsf{T}}

\newcommand{\vC}{\vect{C}}

\newcommand{\vG}{\vect{G}}

\newcommand{\vL}{\vect{L}}

\newcommand{\vP}{\vect{P}}

\newcommand{\vb}{\vect{b}}
\newcommand{\vc}{\vect{c}}

\newcommand{\vl}{\vect{l}}

\newcommand{\vs}{\vect{s}}

\newcommand{\vmu}{\vectg{\mu}}

\newcommand{\vSigma}{\vectg{\Sigma}}

\newcommand{\cD}{\mathcal{D}}

\newcommand{\cG}{\mathcal{G}}

%% file: sections/Introduction.tex
\section{INTRODUCTION}
While visual object detectors have significantly advanced over the past years, their application in \emph{open-set} conditions remains an unsolved challenge \cite{miller2018dropout, dhamija2020overlooked}. In open-set conditions, an object detector can encounter object classes that were not present in the training dataset (\emph{unknown} object classes)\cite{scheirer2012toward}. Even state-of-the-art detectors heavily degrade in performance in open-set conditions \cite{dhamija2020overlooked}, as they tend to misclassify unknown objects with high confidence as one of the detector's \emph{known} training classes \cite{miller2018dropout, dhamija2020overlooked}. This raises serious concerns about the safety of deploying object detectors in open-set environments, particularly in applications where perception failures can have severe consequences \cite{sunderhauf2018limits}, such as autonomous vehicles, domestic and healthcare service robots, and human-robot collaboration. These applications are often open-set, with complex, evolving environments that cannot be fully represented by a training dataset. 

\begin{figure}[t]
    \centering
    \includegraphics[width=.49\textwidth]{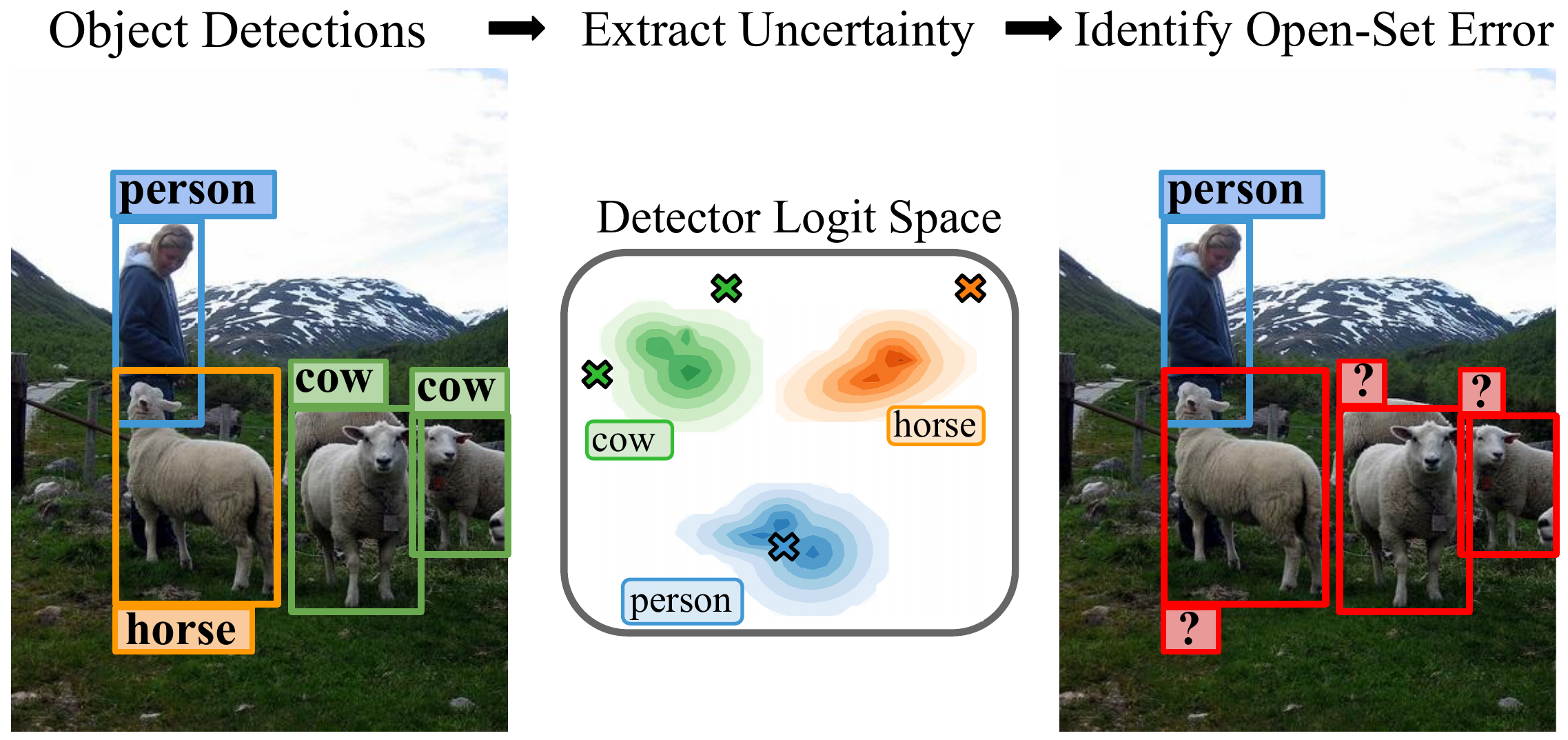}
    \caption{Object detectors make open-set errors when they mistake previously unseen (unknown) objects as known training classes with high confidence -- Faster R-CNN has mistaken sheep (unknown) for known classes cow and horse (with confidences 98\%, 93\% and 90\%). Our proposed \ours{} extracts uncertainty from the detector's logit space by modelling known classes with Gaussian Mixture Models and measuring detection likelihoods. We train with an Anchor loss term to facilitate this modelling. This uncertainty can be used to distinguish between correct detections and open-set errors.}  
    
    \label{fig:bigpic}
\end{figure}

A promising approach to mitigate open-set errors is to analyse the detector's \emph{epistemic uncertainty}, which is uncertainty caused by a lack of knowledge \cite{der2009aleatory}. Object detections with high epistemic uncertainty may indicate open-set errors~\cite{miller2018dropout, miller2019evaluating}, allowing the detector to identify and handle such detections according to the requirements of the application. 

Current techniques for extracting epistemic uncertainty in deep learning models rely heavily on sampling-based techniques such as Monte Carlo (MC) Dropout \cite{gal15dropout} or Deep Ensembles \cite{lakshminarayanan2016simple}. While these techniques have shown promise in the context of open-set object detection \cite{miller2018dropout, miller2019evaluating}, they are computationally expensive, as they require multiple inference passes per image. 
As a result, a standard object detector typically operating at around 30 frames per second can be slowed down to 5 frames per second with these sampling-based techniques.

In the context of robotics and autonomous systems, which are constrained by real-time requirements, a different approach is needed.

We propose \ours{}, an approach for measuring epistemic uncertainty in object detectors that requires only a single inference pass -- thus adding no significant computational overhead to the object detector. We achieve this by adding an Anchor loss term \cite{miller2021class} to the loss function of a detector during training. As we show in our ablation study, this new loss term facilitates the formation of a structured logit space that can be modelled well by a set of class-specific Gaussian Mixture Models (GMMs). During deployment, we can extract epistemic uncertainty for individual detections directly from the set of GMMs via the detection's maximum log-likelihood. \ours{} identifies open-set errors with higher reliability compared to previously proposed methods such as Deep Ensembles and MC Dropout. We visualise this approach in Fig.~\ref{fig:bigpic}. Our method performs especially well when a low rate of open-set errors is of importance -- the operating point of many safety critical applications.

In summary, our paper makes the following contributions:
\begin{enumerate}
    \item We introduce \ours{}, a detector-agnostic method for extracting epistemic uncertainty from object detectors, with minimal added computation.
    \item We show that an Anchor loss term can be added to existing detector loss functions (including cross-entropy loss and focal loss) to facilitate the emergence of a structured logit space.
    \item We show that class-specific Gaussian Mixture Models can be used to model the distribution of known object classes in the detector's logit space, allowing for multiple clusters per class and complex cluster shapes, and supporting identification of open-set errors.
    \item We propose a methodology for adapting existing object detection datasets to support open-set evaluation.
\end{enumerate}

%% file: sections/RelatedWork.tex
\section{RELATED WORK}
\subsection{Uncertainty estimation in object detection}
Visual object detectors localise and classify objects in an image, with a bounding box to describe object location and a class label to describe the object's semantic category. Recently, the task of \emph{probabilistic} object detection was proposed to to include uncertainty estimation \cite{hall2020probabilistic}, where each detection additionally contains a spatial and semantic uncertainty. While a number of recent works have explored spatial uncertainty estimation \cite{he2019bounding, choi2019gaussian, lee2020localization, kraus2019uncertainty, harakeh2020bayesod, phan2018calibrating, he2020deep, harakeh2021estimating} or uncertainty estimation in 3D-LiDAR object detection \cite{feng2018towards, pan2020towards, wang2020inferring, meyer2019lasernet, feng2020leveraging, le2018uncertainty, feng2019can}, we focus on semantic uncertainty estimation for image-based object detection. We refer the reader to \cite{feng2020review} for a review on probabilistic object detection techniques.

Kendall et al. \cite{kendall2017uncertainties} identified epistemic and aleatoric uncertainty as especially relevant for deep networks. Aleatoric uncertainty is due to noise or randomness present in the data (e.g. sensor noise or object occlusions)~\cite{der2009aleatory, kendall2017uncertainties}. Epistemic uncertainty is uncertainty in a model's parameters due to lack of knowledge or data~\cite{der2009aleatory}, and can be used to identify inputs not represented in the model's training data -- epistemic uncertainty is required for identifying open-set errors \cite{kendall2017uncertainties}. 

Object detectors that estimate epistemic uncertainty in the semantic output~\cite{miller2018dropout, miller2019evaluating, harakeh2020bayesod, kraus2019uncertainty, miller2019benchmarking} utilise MC Dropout \cite{gal15dropout} or Deep Ensembles \cite{lakshminarayanan2016simple}. Both methods are sampling-based techniques \cite{miller2019evaluating}, relying on testing an input multiple times and combining the results to obtain an uncertainty estimate. MC Dropout was proposed to approximate a Bayesian Neural Network \cite{gal15dropout}, performing inference several times while dropout is enabled. Deep Ensembles instead performs inference with several distinct models \cite{lakshminarayanan2016simple}. In addition to using MC Dropout, Harakeh et al. \cite{harakeh2020bayesod} proposed a Bayesian post-processing method to replace non-maximum suppression (NMS), showing this improved the uncertainty estimation \cite{harakeh2020bayesod}. In contrast to~\cite{miller2018dropout, miller2019benchmarking, miller2019evaluating, kraus2019uncertainty, harakeh2020bayesod}, we propose a technique for estimating epistemic uncertainty that requires only a single inference pass, adding minimal computation.

\subsection{Open-set classification and object detection}
Standard classifiers and detectors are trained and evaluated in a \emph{closed-set} manner: they train and test on a single set of `known' object classes~\cite{scheirer2012toward}. However, prior knowledge of all possible object classes is not possible for some applications, including robotics~\cite{sunderhauf2018limits}. To address this,~\cite{scheirer2012toward} introduced \emph{open-set} recognition, where a network can also encounter previously unseen, `unknown', classes during the testing phase.

In open-set classification, epistemic uncertainty has been obtained by measuring distance in a classifier's logit space~\cite{bendale2016towards, yoshihashi2019classification, ge2017generative, miller2021class}. This approach assumes that known object classes form clusters and unknown classes will be distinct from these clusters -- an input's distance to each class cluster represents uncertainty. In previous work, we showed that training a classifier with a clustering loss improves distance-based uncertainty for open-set classification~\cite{miller2021class}. This technique has not been explored in object detection, and assumed simple structures in the logit space, with one spherical cluster per class.

Despite a large body of work in open-set classification (see this survey~\cite{Geng2020RecentAI}), only a few works explore the more complex task of open-set object detection~\cite{miller2018dropout, miller2019benchmarking, dhamija2020overlooked}. In previous work, we used MC Dropout to extract epistemic uncertainty from a Single Shot MultiBox Detector for open-set conditions~\cite{miller2018dropout, miller2019evaluating}. Using MC Dropout with a Faster R-CNN detector, \cite{dhamija2020overlooked} conversely found MC Dropout decremented performance in an open-set environment

Dhamija et al. \cite{dhamija2020overlooked} identified a limitation with existing evaluation protocols for open-set object detection. Object detection datasets include a `background' class, which features non-target objects, surfaces and scenery, and represent the data the detector learns to ignore. Current open-set detection evaluation protocols \cite{miller2019evaluating, dhamija2020overlooked} do not distinguish between `known unknown' detections, where the detector encountered the object as part of the background class, and true `unknown' open-set detections. We address this by proposing a method for adapting existing detection datasets to allow for the explicit evaluation of open-set errors.

%% file: sections/Method.tex
\section{OUR APPROACH}
\label{sec:method}
We propose \ours{}, a novel detector-agnostic method for extracting epistemic semantic uncertainty from an object detector. \ours{} achieves this without adding the computational overhead associated with sampling-based methods such as Deep Ensembles~\cite{miller2019benchmarking} or MC Dropout~\cite{miller2018dropout, miller2019evaluating, harakeh2020bayesod}. 

Our core idea is to train an object detector to produce a logit space with known class regions that can be modelled well by a set of class-specific Gaussian Mixture Models (GMMs). There are three elements to our proposed approach which we will explain in detail in the following sections:
\begin{enumerate}[label=(\Alph*)]
    \item During training, we use an Anchor loss term to facilitate learning a structured logit space.
    \item After training, we model `known' regions of the logit space with class-specific Gaussian Mixture Models.
    \item During deployment, we estimate epistemic semantic uncertainty for inputs using the log-likelihood of belonging to `known' regions of the detector's logit space.
\end{enumerate}

\subsection{Training Object Detectors with Structured Logit Spaces}
\subsubsection{Problem Setup}
The detector is trained to localise and classify objects belonging to a set of $N$ known classes in the training dataset. The detector outputs detections $D$ containing bounding box coordinates $\vb$ and class logit vectors $ \vl = (l_1, \dots, l_N)\T$. Typically, a softmax or sigmoid function is applied to normalise $\vs$ to class confidence scores $\vs$ between 0 and 1. We refer to the $N$-dimensional space that a logit vector exists within as the \emph{logit space}.

Existing object detectors use classification losses $\mathcal{L}_\text{cls}$, such as cross-entropy loss or focal loss \cite{lin2017focal}, to train detectors to predict a high positive logit for the correct object class and a low negative logit for all other known classes. While this allows for good performance in closed-set conditions, we show in Section \ref{sec:ablation} that a more constrained structure is required when measuring epistemic uncertainty by modelling the detector logit space.

\subsubsection{Our Approach}
To learn a structured logit space that enables modelling with Gaussian Mixture Models, we train a detected object's logit vector $\vl$ to cluster around class centre points $\vC = (\vc_1, \dots, \vc_N)$ in the logit space. These centre points are fixed during training and have a positive magnitude $\alpha$ in their respective known class dimension and a negative  magnitude $\alpha$ in all others, so that
\begin{equation}
    \vc_1 = (\alpha,-\alpha,\dots,-\alpha)\T, \;\; \dots \;\; \vc_N = (-\alpha, -\alpha, \dots, \alpha)\T.
\end{equation}

We use an Anchor loss term \cite{miller2021class} to minimise the Euclidean distance between $\vl$ and $\vc_y$, given the object belongs to the known class $y$: 

\begin{equation}
    \mathcal{L}_\text{A}(\vl, y) =\| \vl - \vc_y \|_2.
\end{equation}

Given our placement of $\vC$, the Anchor loss term $\mathcal{L}_\text{A}$ has a similar training objective to $\mathcal{L}_\text{cls}$ -- only with a more restrictive requirement for where known classes should map to in the logit space. This allows us to combine $\mathcal{L}_\text{A}$ with the detector's existing $\mathcal{L}_\text{cls}$ 

\begin{equation}
    \overline{\mathcal{L}}_\text{cls}(\vl, y) =  \mathcal{L}_\text{cls}(\vl, y) + \lambda \mathcal{L}_\text{A}(\vl, y),
\end{equation}
\noindent
with a parameter $\lambda$ to weight the Anchor loss. The parameter $\lambda$ reduces the magnitude of $\mathcal{L}_\text{A}$ to balance with $\mathcal{L}_\text{cls}$, and weights the restrictiveness of the clustering imposed by $\mathcal{L}_\text{A}$ during training. We detail recommendations for selecting $\lambda$ in Section \ref{sec:implementation}. For any given detector, we then replace $\mathcal{L}_\text{cls}$ with $\overline{\mathcal{L}}_\text{cls}$ during training to learn a more structured logit space with known classes clustering around $\vC$.

\subsection{Modelling an Object Detector's Logit Space}
To model the known regions of the logit space, we define a set of Gaussian Mixture Models
\begin{equation}
   \cG = \{\vG_1, \vG_2, \dots, \vG_N\}, 
\end{equation}
comprising of a Gaussian Mixture Model (GMM) $\vG_i$ for each known class. The parameters of each $\vG_i$ are estimated using Expectation Maximisation \cite{dempster1977maximum} to fit a set of logit vectors $\vL_i$ from the training dataset. $\vL_i$ contains logit vectors of correctly detected objects from known class $i$ in the training dataset. Given the ground-truth object bounding box $\hat{\vb}$ and class label $i$, we create $\vL_i$ by 
\begin{equation}
    \vl \in \vL_i \iff IoU(\hat{\vb},\vb) \ge \theta_\text{iou} \land \vs_{i} \ge \theta_\text{conf}.
\end{equation}

With $\vs_i$ representing the softmax-normalised (or sigmoid) confidence score for the ground-truth class, the detection must correctly localise the object and assign a high confidence to be included in $\vL_i$. We discuss the selection of $\theta_\text{iou}$ and $\theta_\text{conf}$ in Section \ref{sec:implementation}.

For each $\vG_i$, we must specify the number of components required to model $\vL_i$. We wish to select the number of components that allows each $\vG_i$ to distinguish between detections of objects from known class $i$ and objects from unknown classes -- however, we have no prior knowledge or access to the unknown classes. Instead, we use detections of misclassified objects as a proxy for open-set detections -- while misclassified objects do not optimally represent unknown objects, we were able to attain high performance with this approach. Testing on the validation dataset, we find the number of components that can best distinguish between correctly classified object detections and misclassified object detections (using AUROC, see Section \ref{sec:evalMetrics}).

\subsection{Estimating Epistemic Semantic Uncertainty and Rejecting Errors}

For known class $i$, the defined GMM $\vG_i$ consists of $M$ components, where each individual component $j$ has an estimated mean $\vmu_{i,j}$, covariance matrix $\vSigma_{i,j}$ and component weight $\pi_{i,j}$. During testing, given a detected object and its logit vector $\vl$, we can estimate the log-likelihood that $\hat{\vl}$ belongs to the model $\vG_i$ for class $i$ by
\begin{equation}
    \log\left(p\left(\hat{\vl}; \vG_i\right)\right) = \log\left(\sum_{j=1}^{M}\pi_{i,j}\mathcal{N}\left(\vl;\vmu_{i,j}, \vSigma_{i,j}\right)\right).\\
\end{equation}

We obtain a measure of epistemic uncertainty for each known class by computing the log-likelihood of the data $\hat{l}$ for every known class model $\vG_i$

\begin{equation}
    \vP = \left(\log\left(p\left(\vl; \vG_1\right)\right), \dots,  \log\left(p\left(\vl; \vG_N\right)\right)\right).
\end{equation}

A low log-likelihood represents a high uncertainty the detected object belongs to the respective known class. To identify and reject potential open-set detections, we can choose a minimum log-likelihood threshold $\theta_\text{OSE}$ and reject detections that do not meet this threshold for at least one known class. In some instances, the detected class (via the highest confidence score) differs from the class with the highest log-likelihood. This occurs most frequently for erroneous detections, where between $75-97\%$ of detections with this phenomena were errors. This could be used as another indicator to identify and reject incorrect detections, and we leave this for future work.

\section{CREATING AN OPEN-SET OBJECT DETECTION DATASET}
\label{sec:dataset}
We introduce a methodology for converting any existing object detection dataset into an open-set form. We apply it to Pascal VOC~\cite{everingham2010pascal} and COCO~\cite{lin2014microsoft} in Section \ref{sec:evalData} for our evaluation. First, we explain why existing datasets do not allow for open-set evaluation.

\subsection{Background} 
In an object detection setting, object classes can be categorised into 3 distinct sets:
\begin{enumerate}
    \item \textbf{Known} classes $K$ are labelled in the training dataset and the detector is trained to detect them.
    \item \textbf{Known unknown} classes $U_K$ exist in the training dataset but are unlabelled, or labelled as `background'. The detector is trained to ignore these objects.
    \item \textbf{Unknown unknown} classes $U_U$ are not present in the training dataset. The detector has never seen objects of those classes during training and therefore has not learned to ignore them. Unknown unknowns are the cause for open-set errors.
\end{enumerate}

As identified by~\cite{dhamija2020overlooked}, datasets typically used for open-set object detection~\cite{miller2018dropout, miller2019evaluating, dhamija2020overlooked} do not allow distinction between detections of known unknown (background) objects $U_K$ and unknown unknown objects $U_U$, since neither $U_K$ nor $U_U$ are labelled. This poses an issue for open-set evaluation, as detections of $U_K$ and $U_U$ represent different error types -- detections of $U_K$ represent closed-set error whereas detections of $U_U$ represent open-set error. We therefore propose a method for creating datasets that include labelled $U_U$, thus enabling explicit evaluation of open-set error.

\subsection{Method}

Consider an object detection dataset $\cD$, which contains training, validation and test subsets $\{\cD_\text{Train}, \cD_\text{Val}, \cD_\text{Test}\}$, and includes objects from a set of $N$ labelled classes $K$. First, we split $K$ into two distinct sets: labelled known classes $K_K$ and labelled unknown classes $K_U$. We then create new training, validation and test datasets $\widetilde{\cD}_\text{Train}, \widetilde{\cD}_\text{Val}, \widetilde{\cD}_\text{Test}$ by removing all images from the original subsets that contain objects from labelled unknown classes $K_U$. This way, the classes in $K_U$ can act as unknown unknowns $U_U$, as we ensure they are not seen by the detector during training. We also create a new test dataset $\widetilde{\cD}_\text{Test}$ that does not contain objects from $K_U$, using this to approximate closed-set performance -- note that unlabelled unknown classes $U_U$ may still exist in the dataset's background, as is standard for object detection datasets.

To evaluate performance in open-set object detection, we test the detector on the original test dataset $\cD_\text{Test}$, which now contains labelled known objects $K_K$ and labelled unknown objects $K_U$. If the detector detects an object from $K_U$, this is an open-set error, as as $K_U$ represents unknown unknowns $U_U$ and no objects from $K_U$ exist in the training dataset $\widetilde{\cD}_\text{Train}$. In this way, our open-set object detection dataset allows for a distinction between $U_K$ and $U_U$ and an explicit evaluation of open-set error.

\subsection{Discussion} One consideration when splitting $K$ into $K_K$ and $K_U$ is to ensure that the new $\widetilde{\cD}_\text{Train}$ contains a reasonable number of instances of each class in $K_K$ to still enable learning. This may become an issue when two frequently coexisting classes are split between $K_K$ and $K_U$ -- for example, if `handbag' belongs to $K_K$ and `person' belongs to $K_U$. For this reason, when choosing $K_K$ we recommend ensuring that each known class has a ratio of instances  between $\widetilde{\cD}_\text{Train}$ and $\cD_\text{Train}$ that is greater than or equal to the ratio between $K_K$ and $K$. However, in datasets with many related classes this is not always possible for every single class.

%% file: sections/Evaluation.tex
\section{EVALUATION}
\subsection{Open-Set Datasets}
\label{sec:evalData}
As identified by~\cite{dhamija2020overlooked}, existing object detection datasets are unable to allow for explicit evaluation of open-set object detection. For this reason, we use the methodology described in Section \ref{sec:dataset} to adapt two common benchmarking object detection datasets for our open-set (OS) experiments. All methods are tested on the same following datasets:

\noindent
\textbf{Pascal VOC-OS}: We define the first 15 classes as known classes belonging to $K_K$. The remaining 5 classes are unknown classes belonging to $K_U$. We use the VOC2007 train, VOC2012 train, and VOC2007 validation datasets as $\cD_\text{Train}$, the VOC2012 validation dataset as $\cD_\text{Val}$, and the VOC2007 test dataset as $\cD_\text{Test}$.

\noindent
\textbf{COCO-OS}: We define the first 50 classes as known classes belonging to $K_K$. The remaining 30 classes are unknown classes belonging to $K_U$. We split the COCO2017 training dataset into $\cD_\text{Train}$ and $\cD_\text{Val}$ with an 80/20 split, and use the COCO2017 validation dataset as $\cD_\text{Test}$.

\noindent
\textbf{iCubWorld Transformations}\cite{pasquale2019we}: We additionally test on the iCubWorld Transformations dataset \cite{pasquale2019we} -- collected by the iCub humanoid robot, this dataset features a human holding labelled objects at different rotations and scales and was designed to avoid the visual biases present in many computer vision datasets \cite{pasquale2019we}. Unlike the previous datasets, this dataset does not contain enough labelled data for training, and therefore we cannot apply our proposed dataset adaptation methodology. Instead, we propose the following open-set setup: We test on the set of 6000 labelled images containing known classes: bodylotion, book, ringbinder, flower, mug and soda bottle, and unknown classes: wallet, pencilcase, hairclip and sprayer. We only consider detections that detect the labelled object held by the human. We use detectors trained on the entire dataset of COCO, but only consider COCO classes cup, book, potted plant and bottle as known classes -- thus we only fit GMMs to these classes. We evaluate with this dataset as it represents the challenging conditions that may be encountered 
by an embodied agent. 

\subsection{Measuring Performance}
\label{sec:evalMetrics}
\textit{Categorising detections}: We categorise detections as correct $D_c$, closed-set errors $D_\text{CSE}$ or open-set errors $D_\text{OSE}$. A detection is considered as correct $D_c$ if it localises and predicts the class of a labelled known object $K_K$ in an image. A detection is categorised as an open-set error $D_\text{OSE}$ if it localises a labelled unknown object $K_U$ in an image and misclassifies it as a known class. We consider detections to localise labelled objects when they share an IoU of at least 0.5. All other detections -- which may be due to known class misclassifications, duplicate detections or background detections -- are considered as closed-set errors $D_\text{CSE}$ .

We evaluate both the object detection performance and the uncertainty effectiveness (as done in \cite{miller2019evaluating}). Object detection performance assesses performance at correctly detecting $D_c$ objects belonging to our known target classes and producing minimal false positive detections. To assess this, we use:

\noindent
\textbf{Mean Average Precision (mAP):} measures the mean area under the precision-recall curve for each known class. For high recall, the detector must correctly classify and localise all known objects in the test dataset and for high precision, the detector must minimise the number of closed-set errors $D_\text{CSE}$ (and open-set errors $D_\text{OSE}$ when tested on the open-set dataset). A perfect mAP score is 100\%. 

We also assess the effectiveness of the uncertainty estimation for reducing open-set error -- namely, the ability of an uncertainty measure to accept correct detections $D_c$, while rejecting open-set error $D_\text{OSE}$.

\noindent
\textbf{Receiver Operating Characteristic (ROC) curve:} represents the trade-off between true positive rate (TPR) and false positive rate (FPR) when varying an uncertainty threshold $\theta$ for rejecting detections. For our evaluation, TPR represents the proportion of correct detections $D_c$ that are correctly accepted and FPR represents the proportion of open-set errors $D_\text{OSE}$ that are incorrectly accepted -- for clarity, we henceforth refer to FPR as the open-set error rate (OSR). We calculate TPR and OSR as follows:

\begin{equation}
   TPR(\theta) = \frac{|D_c > \theta |}{|D_c|} \;\;\; OSR(\theta) =  \frac{|D_\text{OSE} > \theta |}{|D_\text{OSE}|}.
    \label{eq:ue}
\end{equation}

We summarise a ROC curve using the \textbf{Area Under the ROC Curve (AUROC)}, where a perfect score is 1.

\noindent
\textbf{True Positive Rate at Open-Set Error Rate:} Following the definition of TPR and OSR above, we report TPR at 5\%, 10\% and 20\% OSR. These operating points are at the low-OSR end of the ROC curve and are of particular interest for safety-critical applications where a low rate of open-set errors is important. We additionally report the \textbf{absolute counts} of true positives and open-set errors at these points, as this is also relevant for practical applications. 

\subsection{Comparison Detectors and State-of-the-art Methods}
\label{sec:implementation}
We test with two state-of-the-art object detectors: RetinaNet \cite{lin2017focal}, a one-stage, anchor-box object detector; and Faster R-CNN \cite{ren2015faster}, a two-stage object detector. We use the Faster R-CNN and RetinaNet implementations available at \cite{mmdetection} and \cite{feng2020review} respectively. For both detectors, we use a ResNet-50 and feature pyramid network (FPN) backbone. 
\noindent
\textbf{GMM-Det Implementation:} For the Anchor loss weight ($\lambda$), we initially train with the recommended value of $0.1$ \cite{miller2021class}. If the validation dataset mAP is lower than expected, we recommend lowering $\lambda$ and retraining. We found $\lambda$ values of $0.1$ and $0.05$ to be suitable for Pascal VOC-OS and COCO-OS respectively. When selecting $\theta_\text{iou}$ and $\theta_\text{conf}$, we found that $\theta_\text{iou}$ values between $0.5-0.8$ and $\theta_\text{conf}$ values between $0.2-0.8$ provided consistently high open-set performance (only a $1\%$ variation in open-set AUROC within this range for both detectors on VOC-OS and COCO-OS). We used $\theta_\text{iou}$ and $\theta_\text{conf}$ as $0.6$ and $0.7$ in our experiments. For best open-set performance, $\theta_\text{OSE}$ should be selected from a validation dataset representing the deployment environment. If this dataset cannot be obtained, we recommend misclassified detections of known objects as a proxy for open-set detections and selecting $\theta_\text{OSE}$ for the desired error rate.

We compare to three baselines and state-of-the-art methods for estimating epistemic uncertainty in object detection:

\noindent
\textbf{Standard --} the detector without modifications to training or testing, in its conventional form. Requiring only one test of an image to obtain uncertainty, this method is a baseline for uncertainty estimation. 

\noindent
\textbf{Ensembles --} a Deep Ensemble of five detectors \cite{miller2019benchmarking}, where each detector has been trained with a random initialisation of weights and randomly shuffled data.  We merge samples from each detector with a BSAS clustering approach, and use an IoU threshold of $0.8$ \cite{miller2019evaluating}.

\noindent
\textbf{BayesOD --} utilises MC Dropout to estimate epistemic uncertainty and replaces NMS with a Bayesian post-processing method \cite{harakeh2020bayesod}. We test each image 10 times to obtain MC Dropout samples. MC Dropout can significantly decrement Faster R-CNN performance \cite{miller2019benchmarking, dhamija2020overlooked}, therefore we only evaluate this method with the RetinaNet detector. 

For the above methods, uncertainty is typically approximated as either the maximum known class confidence score \cite{dhamija2020overlooked, miller2018dropout, miller2019evaluating} or the entropy of the class confidence score distribution \cite{harakeh2020bayesod}. We test both, referring to the uncertainties as `score' or `entropy'.

%% file: sections/Results.tex
\section{RESULTS AND DISCUSSION}
\begin{figure}[t!]
    \centering
    \includegraphics[width=.49\textwidth]{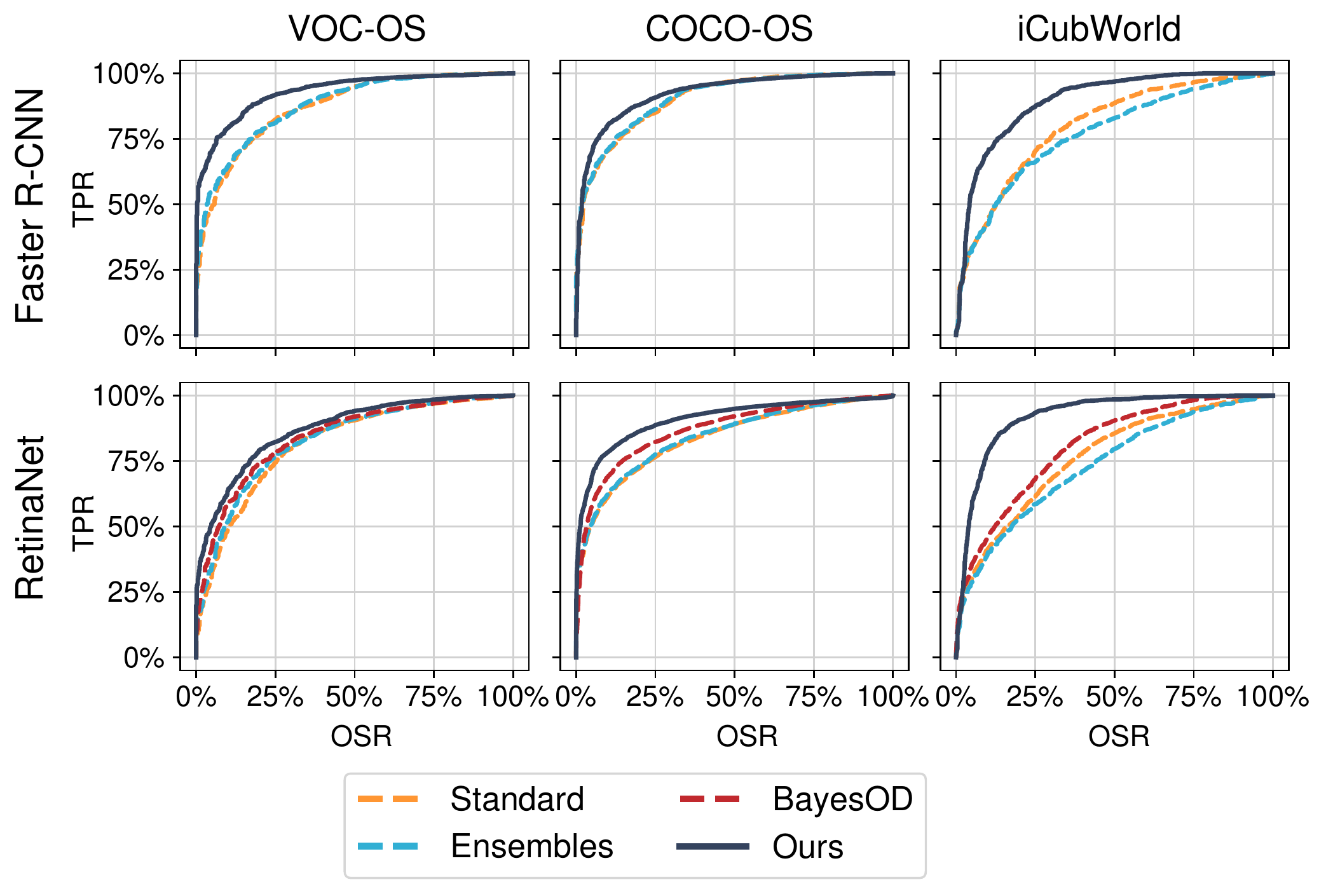}
    \caption{\ours{} produces the best uncertainty performance for accepting correct detections (TP) and rejecting open-set errors (OSE). For comparison methods, we plot the best uncertainty measure (score or entropy) from Table \ref{tab:resultsMain}.}
    \label{fig:roc}
\end{figure}

\begin{table*}[tb]
\centering
\caption{\ours{} (ours) achieves state-of-the-art performance at retaining correct detections while identifying and rejecting open-set error, as measured by AUROC and various true positive rate (TPR) at open-set error rates (OSR).
}
\label{tab:resultsMain}

\setlength\tabcolsep{3.2pt}
\begin{tabular}{@{}lcccc|cccc|cccc@{}}
 \toprule
 
 \textbf{Datasets:}& \multicolumn{4}{c}{\textbf{Pascal VOC-OS}} & \multicolumn{4}{c}{\textbf{COCO-OS}} & \multicolumn{4}{c}{\textbf{iCubWorld Transformations}} \\ 
 \midrule
  & AUROC & \multicolumn{3}{c}{TPR at} & AUROC & \multicolumn{3}{c}{TPR at} & AUROC & \multicolumn{3}{c}{TPR at}\\

  & & 5\%OSR & 10\%OSR & 20\%OSR & & 5\%OSR & 10\%OSR & 20\%OSR & & 5\%OSR & 10\%OSR & 20\%OSR \\

\midrule
\textbf{Faster R-CNN}\\
\midrule
Standard (Score) & 0.861 & 47.2 & 61.7 & 75.6 & 0.876 & 57.1 & 67.4 & 78.0 & 0.773 & 30.5 & 41.6 & 59.0 \\ 

Standard (Entropy) & 0.874 & 48.7 & 62.2 & 76.7 & 0.903 & 59.6 & 70.4 & 81.6 & 0.800 & 33.3 & 43.5 & 62.7 \\ 

Ensembles (Score) & 0.870 & 53.8 & 63.5 & 76.3 & 0.884 & 57.9 & 68.9 & 79.3 & 0.756 & 33.0 & 41.8 & 57.3 \\
Ensembles (Entropy) & 0.879 & 55.0 & 64.3 & 77.8 & 0.906 & 60.0 & 70.8 & 82.4 & 0.771 & 32.1 & 42.2 & 61.6 \\
\ours{} (Ours) & \textbf{0.931} & \textbf{70.2} & \textbf{79.1} & \textbf{89.0} & \textbf{0.924} & \textbf{69.5} & \textbf{80.2} & \textbf{87.9} & \textbf{0.896} & \textbf{54.7} & \textbf{69.6} & \textbf{82.9} \\ 
 \midrule
 \textbf{RetinaNet}\\
\midrule
Standard (Score) & 0.818 & 31.8 & 48.6 & 67.3 & 0.839 & 51.2 & 61.2 & 72.2 & 0.732 & 31.1 & 41.6 & 54.2 \\ 
Standard (Entropy) & 0.814 & 22.2 & 40.0 & 65.8 & 0.801 & 40.3 & 52.7 & 65.8 & 0.766 & 30.2 & 41.4 & 54.9 \\ 

Ensembles (Score) & 0.830 & 34.8 & 51.9 & 71.1 & 0.844 & 51.2 & 62.3 & 73.1 & 0.720 & 32.0 & 40.6 & 53.4\\
Ensembles (Entropy) & 0.797 & 21.7 & 33.3 & 64.6 & 0.778 & 37.7 & 48.6 & 62.2 & 0.736 & 28.7 & 39.3 & 53.2\\
BayesOD (Score) & 0.844 & 41.8 & 58.4 & 73.9 & 0.854 & 55.0 & 64.4 & 75.1 & 0.710 & 29.5 & 37.9 & 50.1 \\ 
BayesOD (Entropy) & 0.782 & 27.5 & 44.3 & 62.2 & 0.871 & 58.0 & 69.0 & 79.0 & 0.808 & 35.5 & 46.0 & 61.7 \\ 
\ours{} (Ours) & \textbf{0.873} & \textbf{53.1} & \textbf{64.2} & \textbf{77.5}  &  \textbf{0.910} & \textbf{68.2} & \textbf{77.4} & \textbf{85.7} & \textbf{0.922} & \textbf{57.3} & \textbf{78.5} & \textbf{90.7} \\ 

\bottomrule
\end{tabular}
\end{table*}

\subsection{Comparison With State-of-the-Art Methods}
\noindent
\textbf{New State-of-the-Art Performance:}
As shown in Table~\ref{tab:resultsMain}, our proposed \ours{} achieves state-of-the-art performance for reliably rejecting open-set error with uncertainty, outperforming all baselines and previously leading methods. This is consistent for both Faster R-CNN and RetinaNet and across all three datasets. For the reported open-set error rates (OSR), the epistemic uncertainty produced by \ours{} is able to preserve more correct (true positive) detections when rejecting open-set error. Across the three datasets at the 5\% OSR, our method improves upon the TPR of the next best method by at least 9.5\% for Faster R-CNN and 10.2\% for RetinaNet. Fig.~\ref{fig:roc} shows this improved performance holds across the complete range of open-set error rates. 

\begin{figure}[tb]
    \centering
    \includegraphics[width=.49\textwidth]{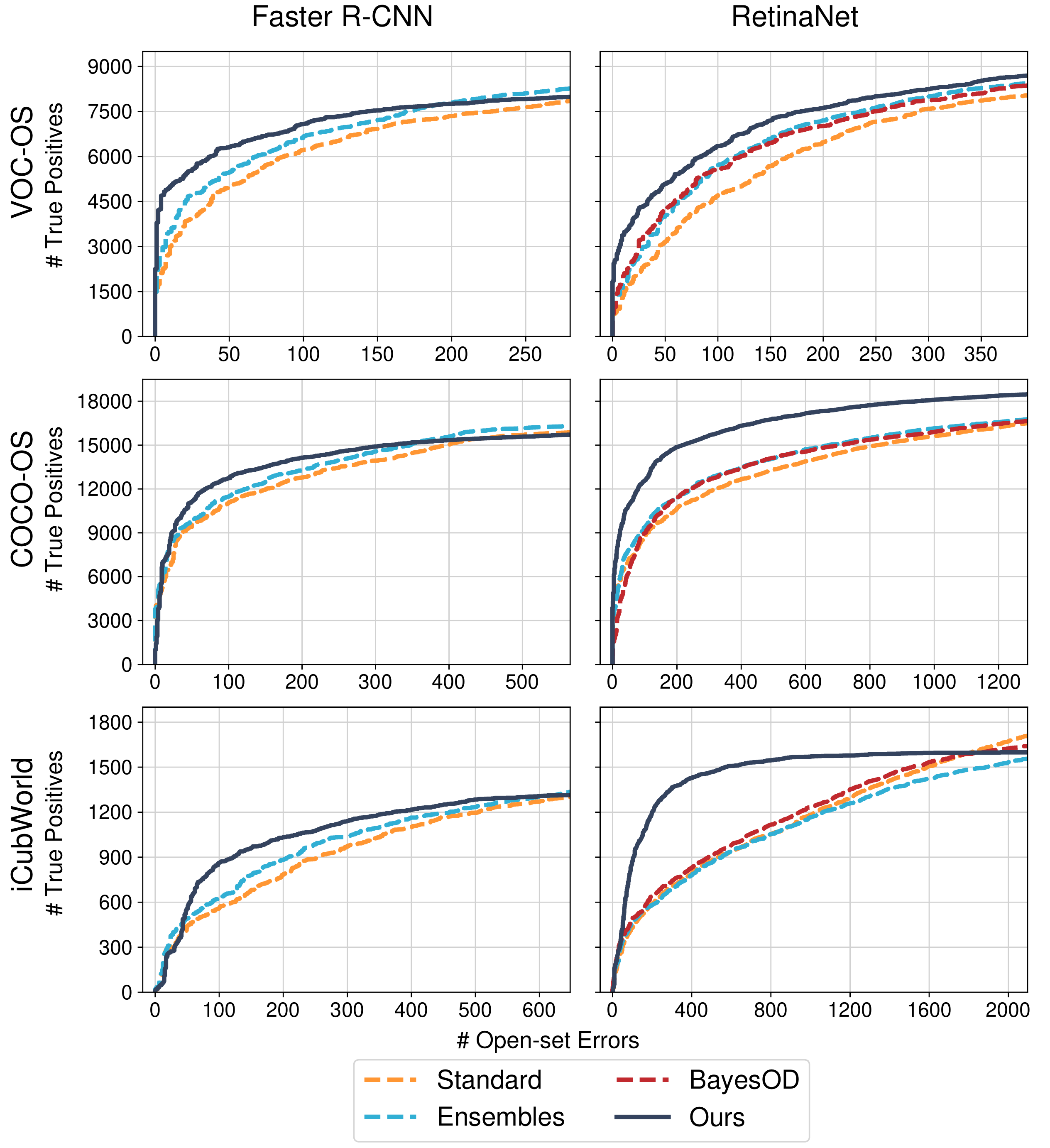}
    \caption{When varying a threshold to reject open-set errors, \ours{} (ours) is able to retain a higher absolute number of correct detections (true positives). Curves are shown until the 50\% OSR operating point for the standard network.}
    \label{fig:absolute}
\end{figure}

Each detector and method produces different numbers of true positives and open-set errors, so we additionally include Fig.~\ref{fig:absolute} to show \emph{absolute} numbers of true positive and open-set detections when varying the uncertainty threshold for rejecting detections. For similar numbers of correct detections, our approach produces fewer open-set errors. In particular, when requiring low numbers of open-set errors, \ours{} is able to achieve noticeably greater numbers of true positive detections. As an example, when requiring at most 100 open-set errors on the COCO-OS dataset, \ours{} retains an extra 1720 and 3797 correct detections over the standard detector for Faster R-CNN and RetinaNet respectively.

\noindent
\textbf{Minimal Computational Overhead:}
As shown in Table ~\ref{tab:FPS}, \ours{} adds only minimal computational overhead to the base detector during inference. Tested on an NVIDIA Titan V, the standard Faster R-CNN and RetinaNet require 35.6ms and 44.1ms respectively. On average, our non-optimised implementation requires an additional 3.5~ms per frame on Faster R-CNN and 6.7~ms on RetinaNet to compute the log-likelihoods from each known class Gaussian Mixture Model. RetinaNet typically produced more detections, and thus required more time for the log-likelihood computation. This is a considerable improvement over the existing state-of-the-art methods, which are sampling-based techniques and introduce an additional 173~ms per frame (Ensembles) or 313~ms per frame (BayesOD) when used in RetinaNet.

\begin{table}
    \centering
    \caption{\ours{} (ours) adds minimal computation time to the standard detector when tested on a NVIDIA Titan V, whereas previous state-of-the-art are 5 to 10 times slower.}
    \label{tab:FPS}
    \begin{tabular}{@{}llrr@{}}
         \toprule
          & & \textbf{Milliseconds/Frame ($\downarrow$)} & \textbf{FPS($\uparrow$)}\\ 
         \midrule
         \multirow{3}{*}{Faster R-CNN} & Standard & \textbf{35.6} & \textbf{28.1}\\
         & \ours{} & 38.9 & 25.7 \\ 
         & Ensembles & 208.3 & 4.8\\ 
         \midrule
         \multirow{4}{*}{RetinaNet} & Standard & \textbf{44.1} & \textbf{22.7} \\ 
         & \ours{} & 50.8 & 19.7 \\ 
         & Ensembles & 217.4 & 4.6 \\ 
         & BayesOD & 357.1 & 2.8 \\ 
         \bottomrule
    \end{tabular}
\end{table}

\noindent
\textbf{Maintaining Object Detection Performance:} \ours{} achieves superior performance for identifying and rejecting open-set error, without impairing the overall performance of the object detector. In Table~\ref{tab:resultsClosedSet}, we show the mAP at IoU 0.5 on the closed-set dataset and the open-set dataset. On the open-set dataset, we threshold the uncertainty to reduce the number of open-set errors to the 20\% OSR of the standard detector. As shown in Table~\ref{tab:resultsClosedSet}, for both the closed-set and open-set datasets, \ours{} RetinaNet improves upon the mAP of the standard detector. We infer that this is due to the addition of the Anchor loss during training, consistent with results found for classification networks \cite{miller2021class}. 

For Faster R-CNN, \ours{} obtains a slightly lower mAP than the standard detector on the closed-set dataset. This is primarily due to a slight drop in recall with our approach. However, on the open-set dataset, \ours{} achieves a higher mAP due to the improved ability to threshold out open-set errors and retain correct detections.

\begin{table}[tb]
\centering
\caption{On the closed-set (CS) dataset and open-set (OS) dataset, \ours{} (ours) maintains a reasonable mAP (at IoU 0.5) when compared to the standard detector.}

\label{tab:resultsClosedSet}

\begin{tabular}{@{}llcc|cc@{}}
 \toprule
 
 \textbf{Datasets:}& & \multicolumn{2}{c}{\textbf{Pascal VOC}} & \multicolumn{2}{c}{\textbf{COCO}} \\ 
 & & \textbf{CS} & \textbf{OS} & \textbf{CS} & \textbf{OS}  \\
 
 \midrule
 \multirow{3}{*}{Faster R-CNN} & Standard & 74.1 & 53.9 & \textbf{52.3} &  41.1\\ 

 & Ensembles \cite{miller2019benchmarking} & \textbf{74.4} & 52.8 & 51.6 & 38.6\\ 
& \ours{} (ours) &  73.0 & \textbf{58.3} & 50.8 & \textbf{41.6}\\ 

 \midrule
 \multirow{4}{*}{RetinaNet} & Standard & 76.4 & 50.1 & 55.0 & 42.9\\ 

& Ensembles \cite{miller2019benchmarking} & 75.8 & 53.4 & 54.4 & 43.0 \\ 

& BayesOD \cite{harakeh2020bayesod} & \textbf{80.2} &   \textbf{59.2} & 53.6 & 42.6\\ 

& \ours{} (ours) & 78.3 & 56.5 & \textbf{55.9} & \textbf{45.7}\\ 

\bottomrule
\end{tabular}

\end{table}

\subsection{Ablation Study}
\label{sec:ablation}
\begin{figure}
    \centering
    \includegraphics[width=.49\textwidth]{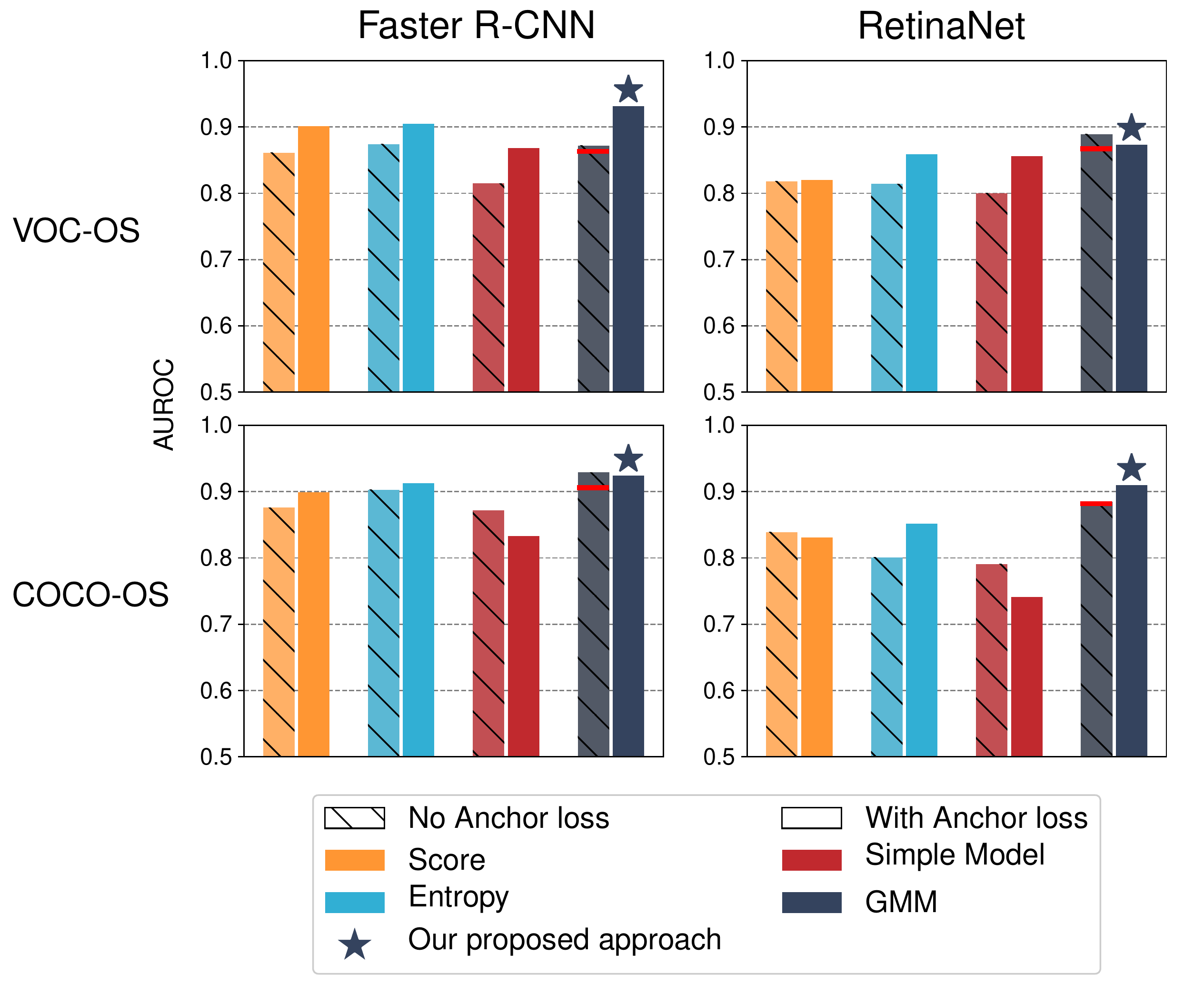}
    \caption{Uncertainty via GMMs outperforms all other uncertainty types, and Anchor loss enables consistent high performance. Performance is measured via AUROC for separating correct detections and open-set errors. A red line indicates the performance of `GMM, No Anchor loss' when constrained to the same number of GMM components as `GMM, With Anchor loss'.}
    \label{fig:ablation}
\end{figure}

We performed an ablation study to quantify the effect of the main components of \ours{} on open-set performance, specifically the influence of Anchor loss on the logit space structure and the capability of Gaussian Mixture Models to approximate this structure. Fig.~\ref{fig:ablation} illustrates the results that we will discuss in the following.

\noindent
\textbf{Unimodal Gaussian vs GMM:}
When modelling the logit space with a \emph{single spherical} Gaussian per class (\cite{miller2021class}), performance drops significantly, as illustrated by the 'Simple Model' versus 'GMM' bars in Fig.~\ref{fig:ablation}. This indicates that the distribution of known classes in the logit space cannot be captured well by a single spherical Gaussian. Although our Anchor loss trains each known class to cluster at a single point, this is unlikely to be achieved when object classes can exhibit objects of different varieties, viewpoints and scales. Interestingly, in~\cite{miller2021class}, we found a single spherical Gaussian sufficient for modelling the logit space for the simpler task of open-set image classification. Further analysing the structure of logit spaces in the context of open-set vision tasks appears to be a worthwhile direction for future research.

\noindent
\textbf{Anchor Loss for Consistent Open-Set Performance and Structured Logit Spaces:} Trained with Anchor loss, the GMM method is able to achieve consistently high open-set performance with $5-6$ mixture components. In some combinations, the detector without Anchor loss achieves slightly higher performance, although this is with the use of a greater number of components -- 14 for Faster R-CNN and 11 for RetinaNet. This highlights the power of the GMM for modelling the logit space, as more components enable the GMM to model more complex structures and distributions. However, when constrained to the same lower number of components, training with Anchor loss consistently achieves higher results.
In addition, training with Anchor loss creates a structured logit space that is more robust to the selection of GMM components. When testing open-set AUROC performance for a range of GMM components between $3-15$, detectors trained without Anchor loss observe up to a $4.1\%$ variation in open-set performance. In contrast, Anchor loss detectors observe at most a $1.6\%$ variation in performance, suggesting logit space structures that are more robust to GMM component selection.

%% file: sections/Conclusion.tex
\section{CONCLUSIONS AND FUTURE WORK}
We have shown that training a detector with an anchor loss term and modelling its logit space with class-specific Gaussian Mixture Models produces epistemic uncertainty that can identify and reject open-set errors. We achieve this result without introducing the significant computational overhead of previous state-of-the-art sampling-based methods, making our work especially relevant for applications of robotics, where open-set conditions are regularly encountered.

The connection of the presented work to active learning would be an interesting avenue to pursue: instead of merely rejecting the identified open-set detections, a robot could use this information to build a new dataset of all unknown objects in its operating environment, acquire ground truth labels from the user, and through continued training, keep adapting its object detection capabilities to its deployment environment.

%% file: main.bbl
\begin{thebibliography}{10}
\providecommand{\url}[1]{#1}
\csname url@samestyle\endcsname
\providecommand{\newblock}{\relax}
\providecommand{\bibinfo}[2]{#2}
\providecommand{\BIBentrySTDinterwordspacing}{\spaceskip=0pt\relax}
\providecommand{\BIBentryALTinterwordstretchfactor}{4}
\providecommand{\BIBentryALTinterwordspacing}{\spaceskip=\fontdimen2\font plus
\BIBentryALTinterwordstretchfactor\fontdimen3\font minus
  \fontdimen4\font\relax}
\providecommand{\BIBforeignlanguage}[2]{{%
\expandafter\ifx\csname l@#1\endcsname\relax
\typeout{** WARNING: IEEEtran.bst: No hyphenation pattern has been}%
\typeout{** loaded for the language `#1'. Using the pattern for}%
\typeout{** the default language instead.}%
\else
\language=\csname l@#1\endcsname
\fi
#2}}
\providecommand{\BIBdecl}{\relax}
\BIBdecl

\bibitem{miller2018dropout}
D.~Miller, L.~Nicholson, F.~Dayoub, and N.~S{\"u}nderhauf, ``Dropout sampling
  for robust object detection in open-set conditions,'' in \emph{IEEE
  International Conference on Robotics and Automation}, 2018, pp. 1--7.

\bibitem{dhamija2020overlooked}
A.~Dhamija, M.~Gunther, J.~Ventura, and T.~Boult, ``The overlooked elephant of
  object detection: Open set,'' in \emph{The IEEE Winter Conference on
  Applications of Computer Vision}, 2020, pp. 1021--1030.

\bibitem{scheirer2012toward}
W.~J. Scheirer, A.~de~Rezende~Rocha, A.~Sapkota, and T.~E. Boult, ``Toward open
  set recognition,'' \emph{IEEE transactions on pattern analysis and machine
  intelligence}, vol.~35, no.~7, pp. 1757--1772, 2012.

\bibitem{sunderhauf2018limits}
N.~S{\"u}nderhauf, O.~Brock, W.~Scheirer, R.~Hadsell, D.~Fox, J.~Leitner,
  B.~Upcroft, P.~Abbeel, W.~Burgard, M.~Milford \emph{et~al.}, ``The limits and
  potentials of deep learning for robotics,'' \emph{The International Journal
  of Robotics Research}, vol.~37, no. 4-5, pp. 405--420, 2018.

\bibitem{der2009aleatory}
A.~Der~Kiureghian and O.~Ditlevsen, ``Aleatory or epistemic? does it matter?''
  \emph{Structural safety}, vol.~31, no.~2, pp. 105--112, 2009.

\bibitem{miller2019evaluating}
D.~Miller, F.~Dayoub, M.~Milford, and N.~S{\"u}nderhauf, ``Evaluating merging
  strategies for sampling-based uncertainty techniques in object detection,''
  in \emph{International Conference on Robotics and Automation}.\hskip 1em plus
  0.5em minus 0.4em\relax IEEE, 2019, pp. 2348--2354.

\bibitem{gal15dropout}
Y.~Gal and Z.~Ghahramani, ``Dropout as a bayesian approximation: Representing
  model uncertainty in deep learning,'' in \emph{Proceedings of The 33rd
  International Conference on Machine Learning}, 2015.

\bibitem{lakshminarayanan2016simple}
B.~Lakshminarayanan, A.~Pritzel, and C.~Blundell, ``Simple and scalable
  predictive uncertainty estimation using deep ensembles,'' in
  \emph{Proceedings of the 31st Conference on Neural Information Processing
  Systems}, 2017, p. 6405–6416.

\bibitem{miller2021class}
D.~Miller, N.~S\"underhauf, M.~Milford, and F.~Dayoub, ``Class anchor
  clustering: A loss for distance-based open set recognition,'' in
  \emph{Proceedings of the IEEE/CVF Winter Conference on Applications of
  Computer Vision}, 2021, pp. 3570--3578.

\bibitem{hall2020probabilistic}
D.~Hall, F.~Dayoub, J.~Skinner, H.~Zhang, D.~Miller, P.~Corke, G.~Carneiro,
  A.~Angelova, and N.~S{\"u}nderhauf, ``Probabilistic object detection:
  Definition and evaluation,'' in \emph{Proceedings of the IEEE/CVF Winter
  Conference on Applications of Computer Vision}, 2020.

\bibitem{he2019bounding}
Y.~He, C.~Zhu, J.~Wang, M.~Savvides, and X.~Zhang, ``Bounding box regression
  with uncertainty for accurate object detection,'' in \emph{Proceedings of the
  IEEE Conference on Computer Vision and Pattern Recognition}, 2019, pp.
  2888--2897.

\bibitem{choi2019gaussian}
J.~Choi, D.~Chun, H.~Kim, and H.-J. Lee, ``Gaussian yolov3: An accurate and
  fast object detector using localization uncertainty for autonomous driving,''
  in \emph{Proceedings of the IEEE International Conference on Computer
  Vision}, 2019, pp. 502--511.

\bibitem{lee2020localization}
Y.~Lee, J.-w. Hwang, H.-I. Kim, K.~Yun, and J.~Park, ``Localization uncertainty
  estimation for anchor-free object detection,'' \emph{arXiv preprint
  arXiv:2006.15607}, 2020.

\bibitem{kraus2019uncertainty}
F.~Kraus and K.~Dietmayer, ``Uncertainty estimation in one-stage object
  detection,'' in \emph{2019 IEEE Intelligent Transportation Systems Conference
  (ITSC)}.\hskip 1em plus 0.5em minus 0.4em\relax IEEE, 2019, pp. 53--60.

\bibitem{harakeh2020bayesod}
A.~Harakeh, M.~Smart, and S.~L. Waslander, ``Bayesod: A bayesian approach for
  uncertainty estimation in deep object detectors,'' in \emph{IEEE
  International Conference on Robotics and Automation}, 2020.

\bibitem{phan2018calibrating}
B.~Phan, R.~Salay, K.~Czarnecki, V.~Abdelzad, T.~Denouden, and S.~Vernekar,
  ``Calibrating uncertainties in object localization task,'' \emph{arXiv
  preprint arXiv:1811.11210}, 2018.

\bibitem{he2020deep}
Y.~He and J.~Wang, ``Deep mixture density network for probabilistic object
  detection,'' in \emph{Proceedings of the International Conference on
  Intelligent Robots and Systems}, 2020.

\bibitem{harakeh2021estimating}
A.~Harakeh and S.~L. Waslander, ``Estimating and evaluating regression
  predictive uncertainty in deep object detectors,'' \emph{arXiv preprint
  arXiv:2101.05036}, 2021.

\bibitem{feng2018towards}
D.~Feng, L.~Rosenbaum, and K.~Dietmayer, ``Towards safe autonomous driving:
  Capture uncertainty in the deep neural network for lidar 3d vehicle
  detection,'' in \emph{2018 21st International Conference on Intelligent
  Transportation Systems (ITSC)}.\hskip 1em plus 0.5em minus 0.4em\relax IEEE,
  2018, pp. 3266--3273.

\bibitem{pan2020towards}
H.~Pan, Z.~Wang, W.~Zhan, and M.~Tomizuka, ``Towards better performance and
  more explainable uncertainty for 3d object detection of autonomous
  vehicles,'' in \emph{2020 IEEE 23rd International Conference on Intelligent
  Transportation Systems (ITSC)}.\hskip 1em plus 0.5em minus 0.4em\relax IEEE,
  2020, pp. 1--7.

\bibitem{wang2020inferring}
Z.~Wang, D.~Feng, Y.~Zhou, W.~Zhan, L.~Rosenbaum, F.~Timm, K.~Dietmayer, and
  M.~Tomizuka, ``Inferring spatial uncertainty in object detection,''
  \emph{arXiv preprint arXiv:2003.03644}, 2020.

\bibitem{meyer2019lasernet}
G.~P. Meyer, A.~Laddha, E.~Kee, C.~Vallespi-Gonzalez, and C.~K. Wellington,
  ``Lasernet: An efficient probabilistic 3d object detector for autonomous
  driving,'' in \emph{Proceedings of the IEEE Conference on Computer Vision and
  Pattern Recognition}, 2019, pp. 12\,677--12\,686.

\bibitem{feng2020leveraging}
D.~Feng, Y.~Cao, L.~Rosenbaum, F.~Timm, and K.~Dietmayer, ``Leveraging
  uncertainties for deep multi-modal object detection in autonomous driving,''
  \emph{arXiv preprint arXiv:2002.00216}, 2020.

\bibitem{le2018uncertainty}
M.~T. Le, F.~Diehl, T.~Brunner, and A.~Knol, ``Uncertainty estimation for deep
  neural object detectors in safety-critical applications,'' in \emph{2018 21st
  International Conference on Intelligent Transportation Systems (ITSC)}.\hskip
  1em plus 0.5em minus 0.4em\relax IEEE, 2018, pp. 3873--3878.

\bibitem{feng2019can}
D.~Feng, L.~Rosenbaum, C.~Glaeser, F.~Timm, and K.~Dietmayer, ``Can we trust
  you? on calibration of a probabilistic object detector for autonomous
  driving,'' \emph{arXiv preprint arXiv:1909.12358}, 2019.

\bibitem{feng2020review}
D.~Feng, A.~Harakeh, S.~Waslander, and K.~Dietmayer, ``A review and comparative
  study on probabilistic object detection in autonomous driving,'' \emph{arXiv
  preprint arXiv:2011.10671}, 2020.

\bibitem{kendall2017uncertainties}
A.~Kendall and Y.~Gal, ``What uncertainties do we need in bayesian deep
  learning for computer vision?'' in \emph{Proceedings of the 31st Concerence
  on Neural Information Processing Systems}, 2017.

\bibitem{miller2019benchmarking}
D.~Miller, N.~S{\"u}nderhauf, H.~Zhang, D.~Hall, and F.~Dayoub, ``Benchmarking
  sampling-based probabilistic object detectors.'' in \emph{CVPR Workshops},
  vol.~3, 2019.

\bibitem{bendale2016towards}
A.~Bendale and T.~E. Boult, ``Towards open set deep networks,'' in
  \emph{Proceedings of the IEEE Conference on Computer Vision and Pattern
  Recognition}, 2016.

\bibitem{yoshihashi2019classification}
R.~Yoshihashi, W.~Shao, R.~Kawakami, S.~You, M.~Iida, and T.~Naemura,
  ``Classification-reconstruction learning for open-set recognition,'' in
  \emph{Proceedings of the IEEE Conference on Computer Vision and Pattern
  Recognition}, 2019, pp. 4016--4025.

\bibitem{ge2017generative}
S.~D. Zongyuan~Ge and R.~Garnavi, ``Generative openmax for multi-class open set
  classification,'' in \emph{Proceedings of the British Machine Vision
  Conference}, 2017, pp. 42.1--42.12.

\bibitem{Geng2020RecentAI}
C.~Geng, S.-J. Huang, and S.~Chen, ``Recent advances in open set recognition: A
  survey,'' \emph{IEEE transactions on pattern analysis and machine
  intelligence}, 2020.

\bibitem{lin2017focal}
T.-Y. Lin, P.~Goyal, R.~Girshick, K.~He, and P.~Doll{\'a}r, ``Focal loss for
  dense object detection,'' in \emph{Proceedings of the IEEE international
  conference on computer vision}, 2017, pp. 2980--2988.

\bibitem{dempster1977maximum}
A.~P. Dempster, N.~M. Laird, and D.~B. Rubin, ``Maximum likelihood from
  incomplete data via the em algorithm,'' \emph{Journal of the Royal
  Statistical Society: Series B (Methodological)}, vol.~39, no.~1, 1977.

\bibitem{everingham2010pascal}
M.~Everingham, L.~Van~Gool, C.~K. Williams, J.~Winn, and A.~Zisserman, ``The
  pascal visual object classes (voc) challenge,'' \emph{International journal
  of computer vision}, vol.~88, no.~2, pp. 303--338, 2010.

\bibitem{lin2014microsoft}
T.-Y. Lin, M.~Maire, S.~Belongie, J.~Hays, P.~Perona, D.~Ramanan,
  P.~Doll{\'a}r, and C.~L. Zitnick, ``Microsoft coco: Common objects in
  context,'' in \emph{European conference on computer vision}.\hskip 1em plus
  0.5em minus 0.4em\relax Springer, 2014.

\bibitem{pasquale2019we}
G.~Pasquale, C.~Ciliberto, F.~Odone, L.~Rosasco, and L.~Natale, ``Are we done
  with object recognition? the icub robot’s perspective,'' \emph{Robotics and
  Autonomous Systems}, vol. 112, pp. 260--281, 2019.

\bibitem{ren2015faster}
S.~Ren, K.~He, R.~Girshick, and J.~Sun, ``Faster r-cnn: Towards real-time
  object detection with region proposal networks,'' \emph{arXiv preprint
  arXiv:1506.01497}, 2015.

\bibitem{mmdetection}
K.~Chen, J.~Wang, J.~Pang, Y.~Cao, Y.~Xiong, X.~Li, S.~Sun, W.~Feng, Z.~Liu,
  J.~Xu, Z.~Zhang, D.~Cheng, C.~Zhu, T.~Cheng, Q.~Zhao, B.~Li, X.~Lu, R.~Zhu,
  Y.~Wu, J.~Dai, J.~Wang, J.~Shi, W.~Ouyang, C.~C. Loy, and D.~Lin,
  ``{MMDetection}: Open mmlab detection toolbox and benchmark,'' \emph{arXiv
  preprint arXiv:1906.07155}, 2019.

\end{thebibliography}
